\documentclass[10pt,twocolumn,letterpaper]{article}

\usepackage{wacv}
\usepackage{times}
\usepackage{epsfig}
\usepackage{graphicx}
\usepackage{amsmath}
\usepackage{amssymb}

\usepackage{makeidx}
\usepackage{graphicx}
\usepackage{amsmath,amssymb} % define this before the line numbering.
\usepackage{color}
\usepackage[utf8]{inputenc}
\usepackage{times}
\usepackage[english]{babel}
\usepackage{amsmath}
\usepackage{url}
\usepackage{lipsum}
\usepackage{changepage} 
\usepackage{enumitem}

\usepackage{booktabs}
\wacvfinalcopy % *** Uncomment this line for the final submission

\def\wacvPaperID{231} % *** Enter the wacv Paper ID here

\ifwacvfinal\fi
\begin{document}

%%%%%%%%% TITLE
\title{Exploring Hate Speech Detection in Multimodal Publications}

% Authors at the same institution
%\author{First Author \hspace{2cm} Second Author \\
% Authors at different institutions
\author{Raul Gomez$^{1,2}$, Jaume Gibert$^{1}$, Lluis Gomez$^{2}$, Dimosthenis Karatzas$^{2}$\\
$^{1}$Eurecat, Centre Tecnològic de Catalunya, Unitat de Tecnologies Audiovisuals, Barcelona, Spain\\
$^{2}$Computer Vision Center, Universitat Autònoma de Barcelona, Barcelona, Spain\\
{\tt\small{\{raul.gomez,jaume.gibert\}@eurecat.org,  \{lgomez,dimos\}@cvc.uab.es}}
}

\maketitle
\ifwacvfinal\thispagestyle{empty}\fi

%%%%%%%%% ABSTRACT
\begin{abstract}
In this work we target the problem of hate speech detection in multimodal publications formed by a text and an image. We gather and annotate a large scale dataset from Twitter, MMHS150K, and propose different models that jointly analyze textual and visual information for hate speech detection, comparing them with unimodal detection.
We provide quantitative and qualitative results and analyze the challenges of the proposed task. 
We find that, even though images are useful for the hate speech detection task, current multimodal models cannot outperform models analyzing only text. We discuss why and open the field and the dataset for further research.
\end{abstract}

%%%%%%%%% BODY TEXT
\section{Introduction}\label{sec:introduction}

% Explain what hate speech is
Social Media platforms such as Facebook, Twitter or Reddit have empowered individuals' voices and facilitated freedom of expression. However they have also been a breeding ground for hate speech and other types of online harassment. Hate speech is defined in legal literature as speech (or any form of expression) that expresses (or seeks to promote, or has the capacity to increase) hatred against a person or a group of people because of a characteristic they share, or a group to which they belong \cite{Herz2012}. Twitter develops this definition in its hateful conduct policy\footnote{\url{https://help.twitter.com/en/rules-and-policies/hateful-conduct-policy}} as \textit{violence against or directly attack or threaten other people on the basis of race, ethnicity, national origin, sexual orientation, gender, gender identity, religious affiliation, age, disability, or serious disease}. 

% Distinguish Hate Speech from offensive language
%In this work we focus on hate speech detection. It is important to distinguish hate speech from other types of online harassment. The presence of offensive terms does not itself signify hate speech. And hate speech may denigrate or threaten an individual or a group of people without the use of any profanities.
%For example some African Americans often use the term \textit{nigga} in everyday language online and the word \textit{cunt} is often used in not hate speech publications and without a sexist meaning.
In this work we focus on hate speech detection. Due to the inherent complexity of this task, it is important to distinguish hate speech from other types of online harassment. In particular, although it might be offensive to many people, the sole presence of insulting terms does not itself signify or convey hate speech. And, the other way around, hate speech may denigrate or threaten an individual or a group of people without the use of any profanities. People from the african-american community, for example, often use the term \textit{nigga} online, in everyday language, without malicious intentions to refer to folks within their community, and the word \textit{cunt} is often used in non hate speech publications and without any sexist purpose. The goal of this work is not to discuss if racial slur, such as \textit{nigga}, should be pursued. The goal is to distinguish between publications using offensive terms and publications attacking communities, which we call hate speech.

\begin{figure}[t]
	\centering
  \includegraphics[width=0.95\linewidth]{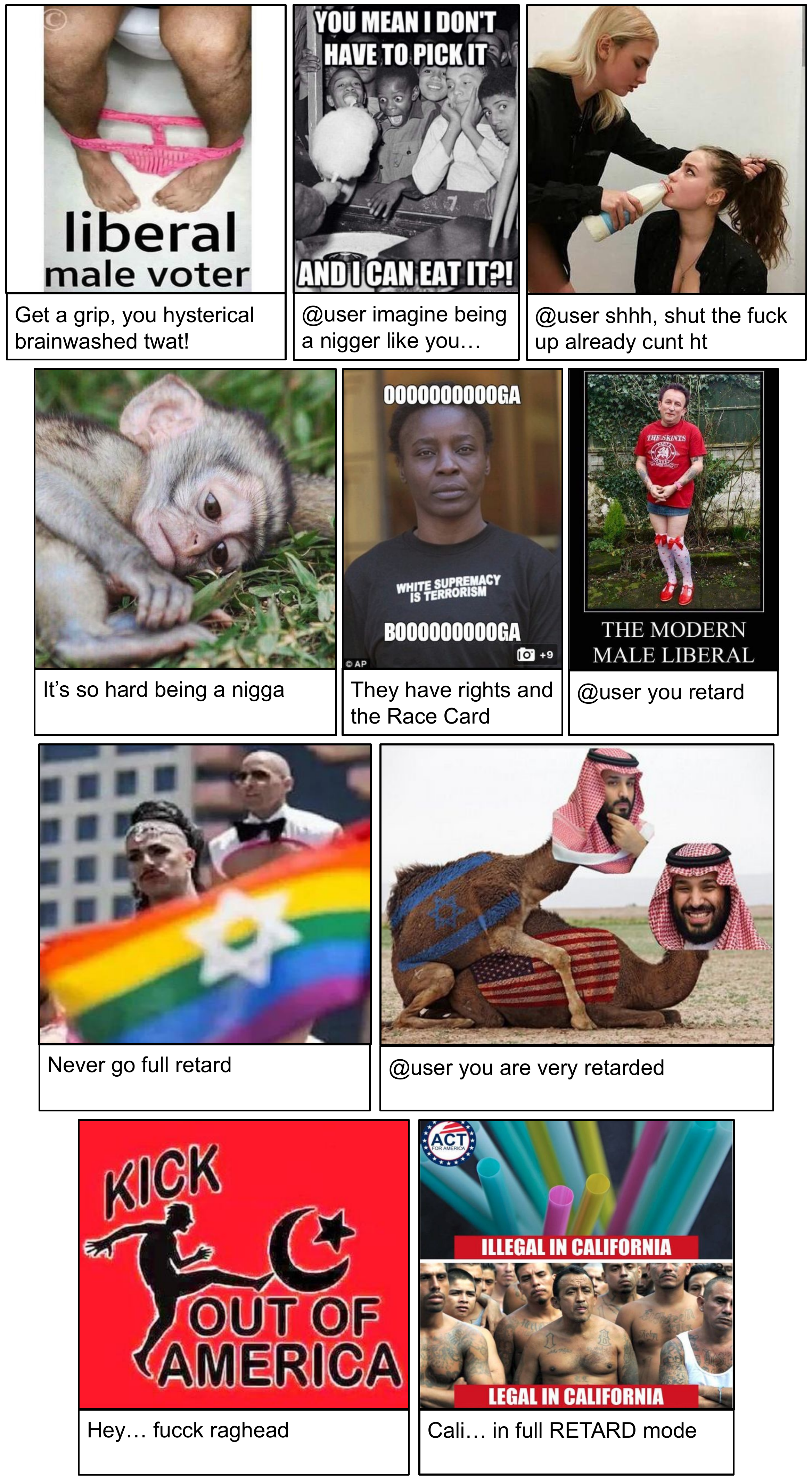}
   \caption{Tweets from MMHS150K where the visual information adds relevant context for the hate speech detection task.}
   \label{fig:visual_hate_examples}
\end{figure}

% Multimodal Hate Speech
Modern social media content usually include images and text. Some of these multimodal publications are only hate speech because of the combination of the text with a certain image. That is because, as we have stated, the presence of offensive terms does not itself signify hate speech, and the presence of hate speech is often determined by the context of a publication. Moreover, users authoring hate speech tend to intentionally construct publications where the text is not enough to determine they are hate speech. This happens especially in Twitter, where multimodal tweets are formed by an image and a short text, which in many cases is not enough to judge them. In those cases, the image might give extra context to make a proper judgement. Fig.~\ref{fig:visual_hate_examples} shows some of such examples in MMHS150K. 

% Explains what we do and our findings
The contributions of this work are as follows:
\begin{itemize}[noitemsep,leftmargin=*]
    \item We propose the novel task of hate speech detection in multimodal publications, collect, annotate and publish a large scale dataset.
    \item We evaluate state of the art multimodal models on this specific task and compare their performance with unimodal detection. Even though images are proved to be useful for hate speech detection, the proposed multimodal models do not outperform unimodal textual models.
    \item We study the challenges of the proposed task, and open the field for future research.
\end{itemize}

\section{Related Work}

% Hate speech detection literature using only text
\subsection{Hate Speech Detection}
The literature on detecting hate speech on online textual publications is extensive. Schmidt and Wiegand \cite{Schmidt} recently provided a good survey of it, where they review the terminology used over time, the features used, the existing datasets and the different approaches. 
However, the field lacks a consistent dataset and evaluation protocol to compare proposed methods.
Saleem \textit{et al.} \cite{Saleem2017} compare different classification methods detecting hate speech in Reddit and other forums.
Wassem and Hovy \cite{WaseemFix} worked on hate speech detection on twitter, published a manually annotated dataset and studied its hate distribution.
Later Wassem \cite{Waseem2016} extended the previous published dataset and compared amateur and expert annotations, concluding that amateur annotators are more likely than expert annotators to label items as hate speech.
Park and Fung \cite{Park} worked on Wassem datasets and proposed a classification method using a CNN over Word2Vec \cite{Mikolov2013} word embeddings, showing also classification results on racism and sexism hate sub-classes.
Davidson \textit{et al.} \cite{Davidson2017} also worked on hate speech detection on twitter, publishing another manually annotated dataset. They test different classifiers such as SVMs and decision trees and provide a performance comparison.
Malmasi and Zampieri \cite{Malmasi2017} worked on Davidson's dataset improving his results using more elaborated features.
ElSherief \textit{et al.} \cite{ElSherief2018} studied hate speech on twitter and selected the most frequent terms in hate tweets based on Hatebase\footnote{\url{https://www.hatebase.org/}}, a hate expression repository. They propose a big hate dataset but it lacks manual annotations, and all the tweets containing certain hate expressions are considered hate speech.
Zhang \textit{et al.} \cite{Zhang2018} recently proposed a more sophisticated approach for hate speech detection, using a CNN and a GRU \cite{Cho} over Word2Vec \cite{Mikolov2013} word embeddings. They show experiments in different datasets outperforming previous methods. Next, we summarize existing hate speech datasets:

\begin{itemize}[noitemsep,leftmargin=*]
    \item RM \cite{Zhang2018}: Formed by $2,435$ tweets discussing Refugees and Muslims, annotated as hate or non-hate.
    \item DT \cite{Davidson2017}: Formed by $24,783$ tweets annotated as hate, offensive language or neither. In our work, offensive language tweets are considered as non-hate.
    \item WZ-LS \cite{Park}: A combination of Wassem datasets \cite{Waseem2016,WaseemFix} labeled as racism, sexism, neither or both that make a total of $18,624$ tweets.
    \item Semi-Supervised \cite{ElSherief2018}: Contains $27,330$ general hate speech Twitter tweets crawled in a semi-supervised manner. 
\end{itemize}

% Hate speech detection literature using also visual information
Although often modern social media publications include images, not too many contributions exist that exploit visual information.
Zhong \textit{et al.} \cite{Zhong} worked on classifying Instagram images as potential cyberbullying targets, exploiting both the image content, the image caption and the comments. However, their visual information processing is limited to the use of features extracted by a pre-trained CNN, the use of which does not achieve any improvement.
Hosseinmardi \textit{et al.} \cite{Hosseinmardi} also address the problem of detecting cyberbullying incidents on Instagram exploiting both textual and image content. But, again, their visual information processing is limited to use the features of a pre-trained CNN, and the improvement when using visual features on cyberbullying classification is only of 0.01\%.

%{\color{red} LG says: In this work we differentiate from previous literature in that: 1: larger dataset, 2: multi-modal, etc... }

\subsection{Visual and Textual Data Fusion}
% Mention some works using multi modal text+image data
A typical task in multimodal visual and textual analysis is to learn an alignment between feature spaces. To do that, usually a CNN and a RNN are trained jointly to learn a joint embedding space from aligned multimodal data. This approach is applied in tasks such as image captioning \cite{Karpathy,Jiang2018} and multimodal image retrieval \cite{DianeLarlus2017,Gomez2018}. On the other hand, instead of explicitly learning an alignment between two spaces, the goal of Visual Question Answering (VQA) is to merge both data modalities in order to decide which answer is correct. This problem requires modeling very precise correlations between the image and the question representations. The VQA task requirements are similar to our hate speech detection problem in multimodal publications, where we have a visual and a textual input and we need to combine both sources of information to understand the global context and make a decision. We thus take inspiration from the VQA literature for the tested models. 
Early VQA methods \cite{Zhou2015} fuse textual and visual information by feature concatenation. Later methods, such as Multimodal Compact Bilinear pooling \cite{Fukui}, utilize bilinear pooling to learn multimodal features. An important limitation of these methods is that the multimodal features are fused in the latter model stage, so the textual and visual relationships are modeled only in the last layers. Another limitation is that the visual features are obtained by representing the output of the CNN as a one dimensional vector, which losses the spatial information of the input images. In a recent work, Gao \textit{et al.} \cite{Gao2018} propose a feature fusion scheme to overcome these limitations. They learn convolution kernels from the textual information --which they call question-guided kernels-- and convolve them with the visual information in an earlier stage to get the multimodal features. Margffoy-Tuay \textit{et al.} \cite{Margffoy-Tuay2018} use a similar approach to combine visual and textual information, but they address a different task: instance segmentation guided by natural language queries. We inspire in these latest feature fusion works to build the models for hate speech detection.

\section{The MMHS150K dataset}
Existing hate speech datasets contain only textual data. Moreover, a reference benchmark does not exists. Most of the published datasets are crawled from Twitter and distributed as tweet IDs but, since Twitter removes reported user accounts, an important amount of their hate tweets is no longer accessible. We create a new manually annotated multimodal hate speech dataset formed by $150,000$ tweets, each one of them containing text and an image. We call the dataset MMHS150K, and made it available online~\footnote{\url{https://gombru.github.io/2019/10/09/MMHS/}}. In this section, we explain the dataset creation steps.

\subsection{Tweets Gathering}
We used the Twitter API to gather real-time tweets from September $2018$ until February $2019$, selecting the ones containing any of the $51$ Hatebase terms that are more common in hate speech tweets, as studied in \cite{ElSherief2018}. We filtered out retweets, tweets containing less than three words and tweets containing porn related terms. From that selection, we kept the ones that included images and downloaded them. 
Twitter applies hate speech filters and other kinds of content control based on its policy, although the supervision is based on users' reports. Therefore, as we are gathering tweets from real-time posting, the content we get has not yet passed any filter. 

\subsection{Textual Image Filtering}
We aim to create a multimodal hate speech database where all the instances contain visual and textual information that we can later process to determine if a tweet is hate speech or not. But a considerable amount of the images of the selected tweets contain only textual information, such as screenshots of other tweets. 
To ensure that all the dataset instances contain both visual and textual information, we remove those tweets. To do that, we use TextFCN \cite{Bazazian2016,Bazazian2017}
%\footnote{\url{https://github.com/gombru/TextFCN}}
, a Fully Convolutional Network that produces a pixel wise text probability map of an image. We set empirical thresholds to discard images that have a substantial total text probability, filtering out $23\%$ of the collected tweets.

\subsection{Annotation}
We annotate the gathered tweets using the crowdsourcing platform Amazon Mechanical Turk. There, we give the workers the definition of hate speech and show some examples to make the task clearer. We then show the tweet text and image and we ask them to classify it in one of $6$ categories: \textit{No attacks to any community}, \textit{racist}, \textit{sexist}, \textit{homophobic}, \textit{religion based attacks} or \textit{attacks to other communities}.
Each one of the $150,000$ tweets is labeled by 3 different workers to palliate discrepancies among workers. 

\begin{figure}[ht]
	\centering
  \includegraphics[width=0.8\linewidth]{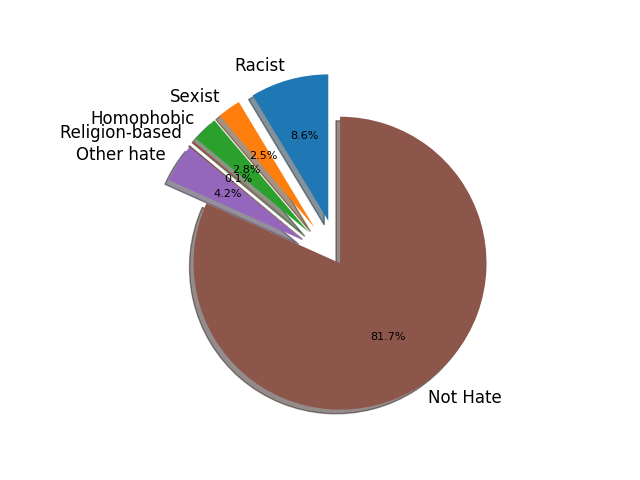}
   \caption{Percentage of tweets per class in MMHS150K.}
   \label{fig:classes_pie}
\end{figure}

\begin{figure*}[t]
	\centering
  \includegraphics[width=0.8\linewidth]{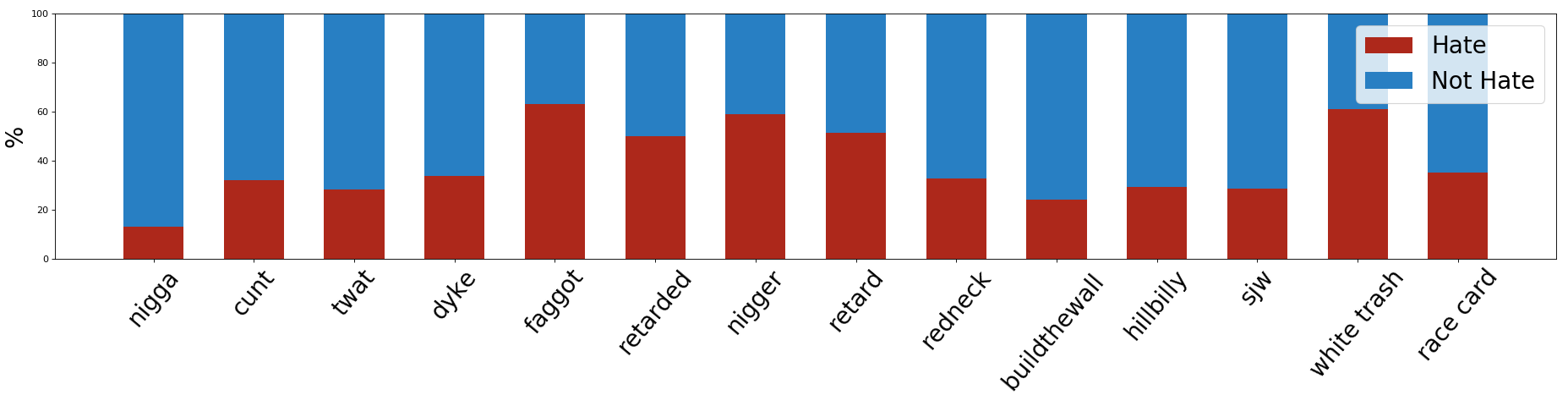}
   \caption{Percentage of hate and not hate tweets for top keywords of MMHS150K.}
   \label{fig:hate_per_word}
\end{figure*}

We received a lot of valuable feedback from the annotators. Most of them had understood the task correctly, but they were worried because of its subjectivity. This is indeed a subjective task, highly dependent on the annotator convictions and sensitivity. However, we expect to get cleaner annotations the more strong the attack is, which are the publications we are more interested on detecting. 
We also detected that several users annotate tweets for hate speech just by spotting slur. As already said previously, just the use of particular words can be offensive to many people, but this is not the task we aim to solve. We have not included in our experiments those hits that were made in less than 3 seconds, understanding that it takes more time to grasp the multimodal context and make a decision.

We do a majority voting between the three annotations to get the tweets category. 
At the end, we obtain $112,845$ not hate tweets and $36,978$ hate tweets. The latest are divided in $11,925$ racist, $3,495$ sexist, $3,870$ homophobic, $163$ religion-based hate and $5,811$ other hate tweets (Fig.~\ref{fig:classes_pie}). In this work, we do not use hate sub-categories, and stick to the hate / not hate split. We separate balanced validation ($5,000$) and test ($10,000$) sets. The remaining tweets are used for training. 

We also experimented using hate scores for each tweet computed given the different votes by the three annotators instead of binary labels. The results did not present significant differences to those shown in the experimental part of this work, but the raw annotations will be published nonetheless for further research.

As far as we know, this dataset is the biggest hate speech dataset to date, and the first multimodal hate speech dataset.
One of its challenges is to distinguish between tweets using the same key offensive words that constitute or not an attack to a community (hate speech). Fig.~\ref{fig:hate_per_word} shows the percentage of hate and not hate tweets of the top keywords.

\section{Methodology}
\subsection{Unimodal Treatment}

\subsubsection{Images.}
All images are resized such that their shortest size has $500$ pixels. During training, online data augmentation is applied as random cropping of $299\times299$ patches and mirroring.
We use a CNN as the image features extractor which is an Imagenet \cite{Deng} pre-trained Google Inception v3 architecture \cite{Szegedy}. The fine-tuning process of the Inception v3 layers aims to modify its weights to extract the features that, combined with the textual information, are optimal for hate speech detection.

\subsubsection{Tweet Text.}
We train a single layer LSTM with a $150$-dimensional hidden state for hate / not hate classification. The input dimensionality is set to $100$ and GloVe \cite{Pennington} embeddings are used as word input representations. Since our dataset is not big enough to train a GloVe word embedding model, we used a pre-trained model that has been trained in two billion tweets\footnote{\url{https://nlp.stanford.edu/projects/glove/}}. This ensures that the model will be able to produce word embeddings for slang and other words typically used in Twitter. To process the tweets text before generating the word embeddings, we use the same pipeline as the model authors, which includes generating symbols to encode Twitter special interactions such as user mentions (@user) or hashtags (\#hashtag).
To encode the tweet text and input it later to multimodal models, we use the LSTM hidden state after processing the last tweet word. Since the LSTM has been trained for hate speech classification, it extracts the most useful information for this task from the text, which is encoded in the hidden state after inputting the last tweet word.

\subsubsection{Image Text.}
The text in the image can also contain important information to decide if a publication is hate speech or not, so we extract it and also input it to our model. To do so, we use Google Vision API Text Detection module \cite{GoogleOCR}. 
We input the tweet text and the text from the image separately to the multimodal models, so it might learn different relations between them and between them and the image. For instance, the model could learn to relate the image text with the area in the image where the text appears, so it could learn to interpret the text in a different way depending on the location where it is written in the image.
The image text is also encoded by the LSTM as the hidden state after processing its last word.

\subsection{Multimodal Architectures}

The objective of this work is to build a hate speech detector that leverages both textual and visual data and detects hate speech publications based on the context given by both data modalities. To study how the multimodal context can boost the performance compared to an unimodal context we evaluate different models: a Feature Concatenation Model (FCM), a Spatial Concatenation Model (SCM) and a Textual Kernels Model (TKM).
All of them are CNN+RNN models with three inputs: the tweet image, the tweet text and the text appearing in the image (if any).

\subsubsection{Feature Concatenation Model (FCM)}
\begin{figure*}[ht]
	\centering
  \includegraphics[width=0.65\linewidth]{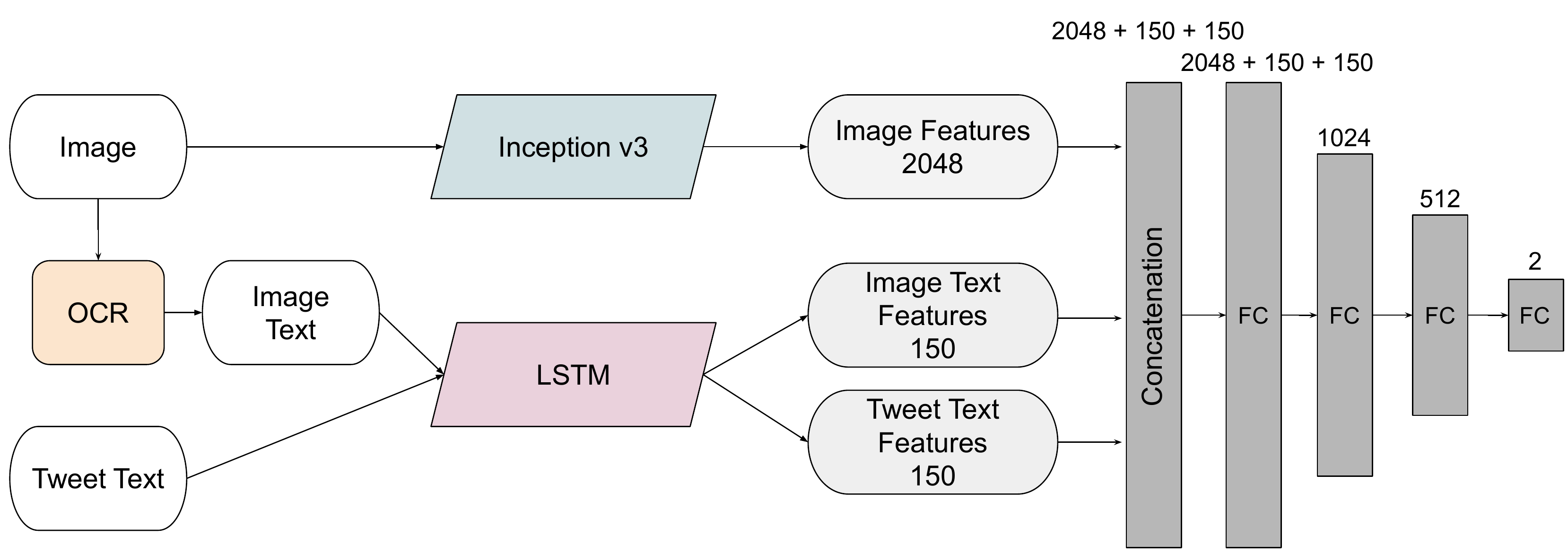}
   \caption{FCM architecture. Image and text representations are concatenated and processed by a set of fully connected layers.}
   \label{fig:model_pipeline}
\end{figure*}

The image is fed to the Inception v3 architecture and the $2048$ dimensional feature vector after the last average pooling layer is used as the visual representation. This vector is then concatenated with the $150$ dimension vectors of the LSTM last word hidden states of the image text and the tweet text, resulting in a $2348$ feature vector. This vector is then processed by three fully connected layers of decreasing dimensionality $(2348, 1024, 512)$ with following corresponding batch normalization and ReLu layers until the dimensions are reduced to two, the number of classes, in the last classification layer.
The FCM architecture is illustrated in Fig.~\ref{fig:model_pipeline}.

\subsubsection{Spatial Concatenation Model (SCM)}
Instead of using the latest feature vector before classification of the Inception v3 as the visual representation, in the SCM we use the $8\times8\times2048$ feature map after the last Inception module. Then we concatenate the $150$ dimension vectors encoding the tweet text and the tweet image text at each spatial location of that feature map. The resulting multimodal feature map is processed by two Inception-E blocks \cite{Szegedy2015}. After that, dropout and average pooling are applied and, as in the FCM model, three fully connected layers are used to reduce the dimensionality until the classification layer.

\subsubsection{Textual Kernels Model (TKM)}
\begin{figure*}[ht]
	\centering
  \includegraphics[width=0.75\linewidth]{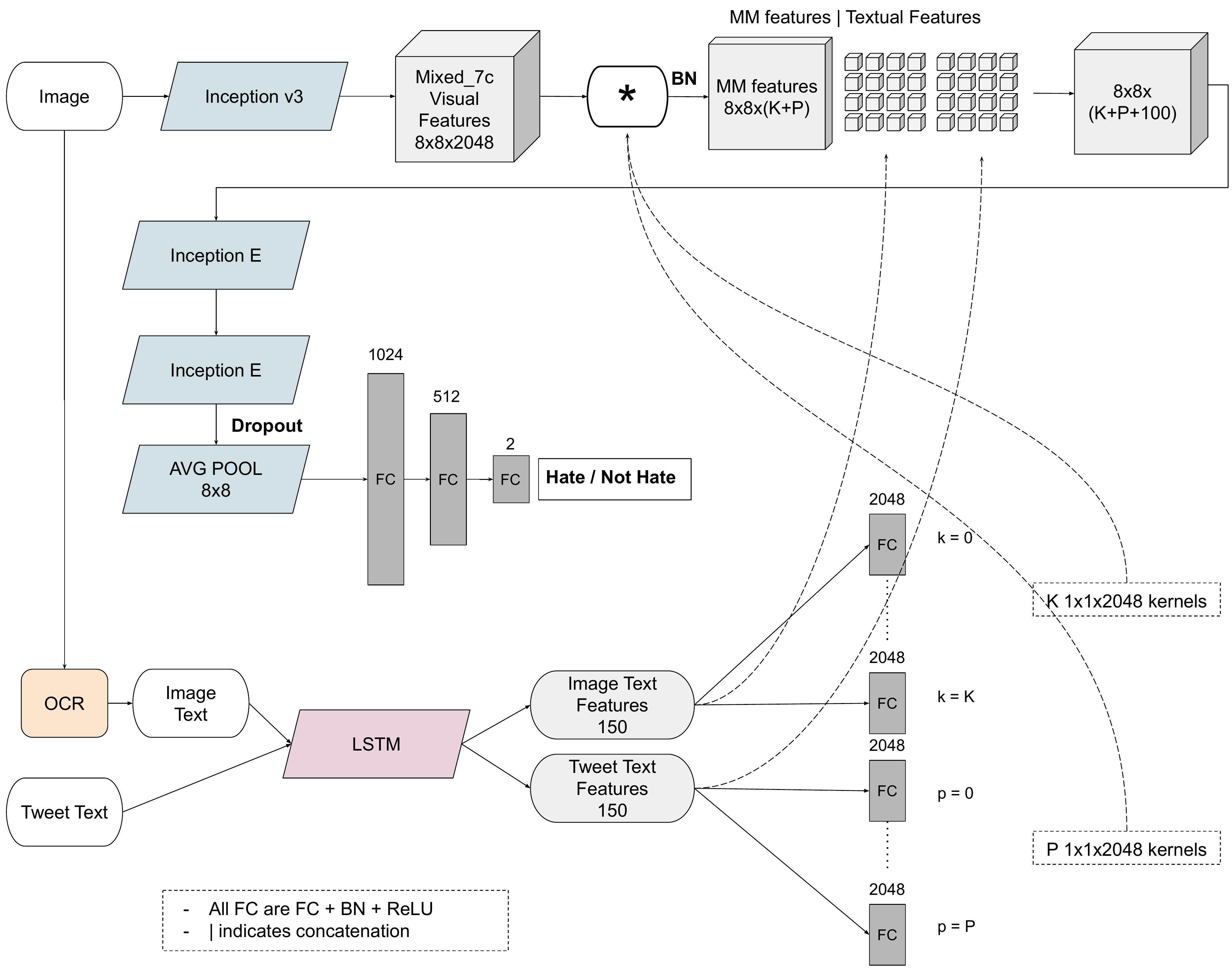}
   \caption{TKM architecture. Textual kernels are learnt from the text representations, and convolved with the image representation.}
   \label{fig:model_pipeline_TKM}
\end{figure*}

The TKM design, inspired by \cite{Gao2018} and \cite{Margffoy-Tuay2018}, aims to capture interactions between the two modalities more expressively than concatenation models. 
As in SCM we use the $8\times8\times2048$ feature map after the last Inception module as the visual representation. From the $150$ dimension vector encoding the tweet text, we learn $K_t$ text dependent kernels using independent fully connected layers that are trained together with the rest of the model. The resulting $K_t$ text dependent kernels will have dimensionality of $1\times1\times2048$.
We do the same with the feature vector encoding the image text, learning $K_{it}$ kernels.
The textual kernels are convolved  with the visual feature map in the channel dimension at each spatial location, resulting in a $8\times8\times(K_i+K_{it})$ multimodal feature map, and batch normalization is applied. Then, as in the SCM, the $150$ dimension vectors encoding the tweet text and the tweet image text are concatenated at each spatial dimension. The rest of the architecture is the same as in SCM: two Inception-E blocks, dropout, average pooling and three fully connected layers until the classification layer. 
The number of tweet textual kernels $K_t$ and tweet image textual kernels $K_it$ is set to $K_t = 10$ and $K_it = 5$. 
The TKM architecture is illustrated in Fig.~\ref{fig:model_pipeline_TKM}.

\subsubsection{Training}
We train the multimodal models with a Cross-Entropy loss with Softmax activations and an ADAM optimizer with an initial learning rate of $1e-4$. Our dataset suffers from a high class imbalance, so we weight the contribution to the loss of the samples to totally compensate for it.
One of the goals of this work is to explore how every one of the inputs contributes to the classification and to prove that the proposed model can learn concurrences between visual and textual data useful to improve the hate speech classification results on multimodal data. To do that we train different models where all or only some inputs are available. When an input is not available, we set it to zeros, and we do the same when an image has no text.

\section{Results}

\begin{figure}[ht]
	\centering
  \includegraphics[width=0.9\linewidth]{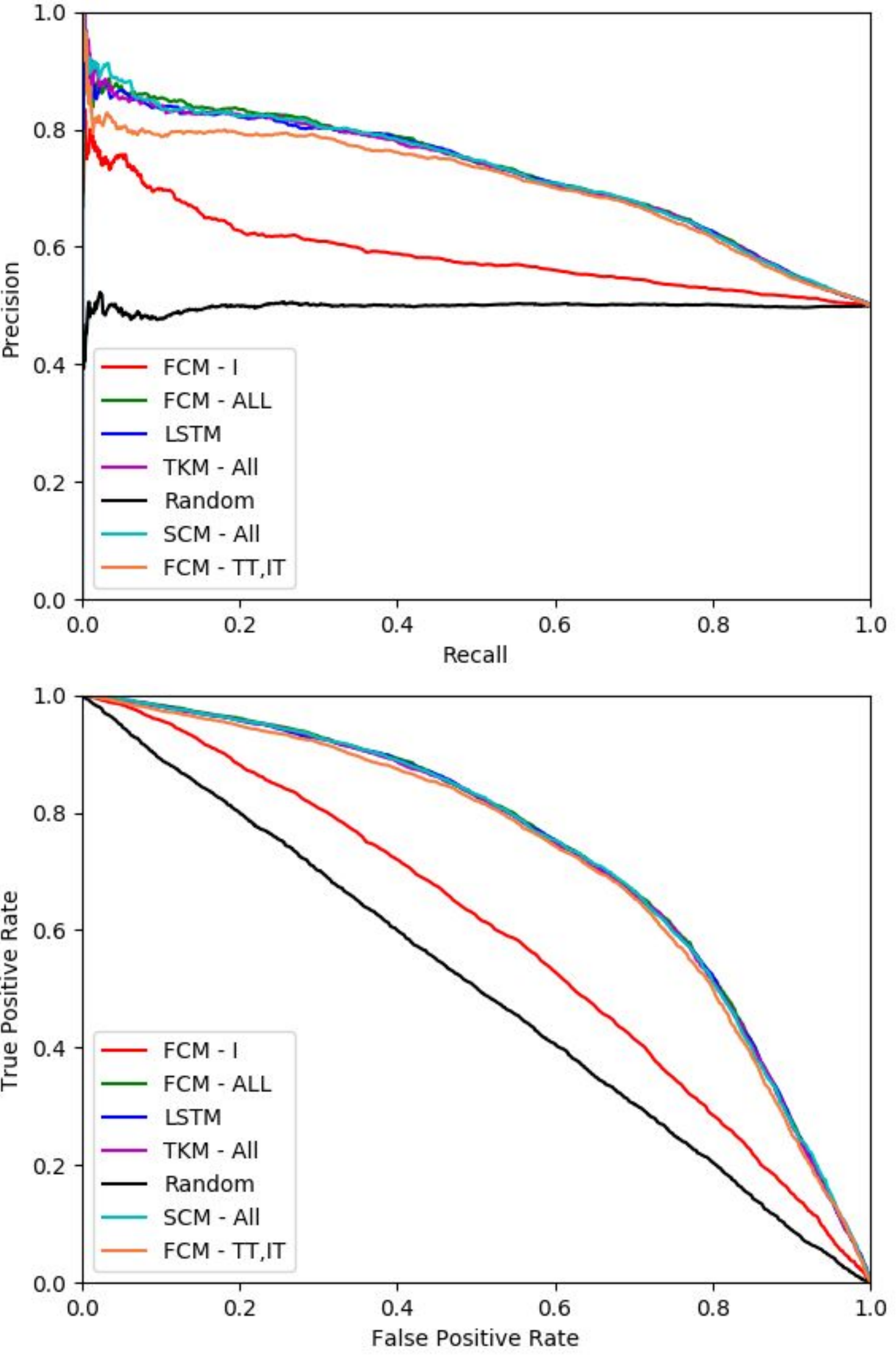}
   \caption{Precision vs Recall (left) and ROC curve (True Positive Rate vs False Positive Rate) (right) plots of the proposed models trained with the different inputs, the LSTM and random scores.}
   \label{fig:plots}
\end{figure}

Table~\ref{table:results} shows the F-score, the Area Under the ROC Curve (AUC) and the mean accuracy (ACC) of the proposed models when different inputs are available. $TT$ refers to the tweet text, $IT$ to the image text and $I$ to the image. It also shows results for the LSTM, for the Davison method proposed in \cite{Davidson2017} trained with MMHS150K, and for random scores.
Fig.~\ref{fig:plots} shows the Precision vs Recall plot and the ROC curve (which plots the True Positive Rate vs the False Positive Rate) of the different models.

\begin{table}[ht]
\centering

%\begin{tabular}{|c|c|c|c|c|}
%\hline
\begin{tabular}{llccccc}
\toprule
\textbf{Model}\hspace*{1em} & \textbf{Inputs}\hspace*{0.5em} & \textbf{F} \hspace*{0.5em} & \textbf{AUC} \hspace*{0.5em} & \textbf{ACC} \\ %\hline
\midrule
Random         & -                                      & 0.666 & 0.499 & 50.2 \\ %\hline
Davison \cite{Davidson2017} & \textit{TT}               & 0.703 & 0.732 & 68.4 \\ %\hline
LSTM           & \textit{TT}                            & 0.703 & 0.732 & 68.3 \\ %\hline
FCM            & \textit{TT}                            & 0.697 & 0.727 & 67.8 \\ %\hline
FCM            & \textit{TT}, \textit{IT}               & 0.697 & 0.722 & 67.9 \\ %\hline
FCM            & \textit{I}                             & 0.667 & 0.589 & 56.8 \\ %\hline
FCM            & \textit{TT}, \textit{IT}, \textit{I}   & 0.704 & 0.734 & 68.4 \\ %\hline
SCM            & \textit{TT}, \textit{IT}, \textit{I}   & 0.702 & 0.732 & 68.5 \\ %\hline
TKM            & \textit{TT}, \textit{IT}, \textit{I}   & 0.701 & 0.731 & 68.2 \\ %\hline 
\bottomrule
\\
\end{tabular}
\caption{Performance of the proposed models, the LSTM and random scores. The \textbf{Inputs} column indicate which inputs are available at training and testing time.}
\label{table:results}
\end{table}

% Optimum results are not 100%
First, notice that given the subjectivity of the task and the discrepancies
%\footnote{this is why we should give values before, because it's a very hard task!}
between annotators, getting optimal scores in the evaluation metrics is virtually impossible. However, a system with relatively low metric scores can still be very useful for hate speech detection in a real application: it will fire on publications for which most annotators agree they are hate, which are often the stronger attacks.
% Our LSTM gets the same as Davison but our objective is multimodal
The proposed LSTM to detect hate speech when only text is available, gets similar results as the method presented in \cite{Davidson2017}, which we trained with MMHS150K and the same splits. 
However, more than substantially advancing the state of the art on hate speech detection in textual publications, our key purpose in this work is to introduce and work on its detection on multimodal publications. We use LSTM because it provides a strong representation of the tweet texts.

\begin{figure}[ht]
	\centering
  \includegraphics[width=0.8\linewidth]{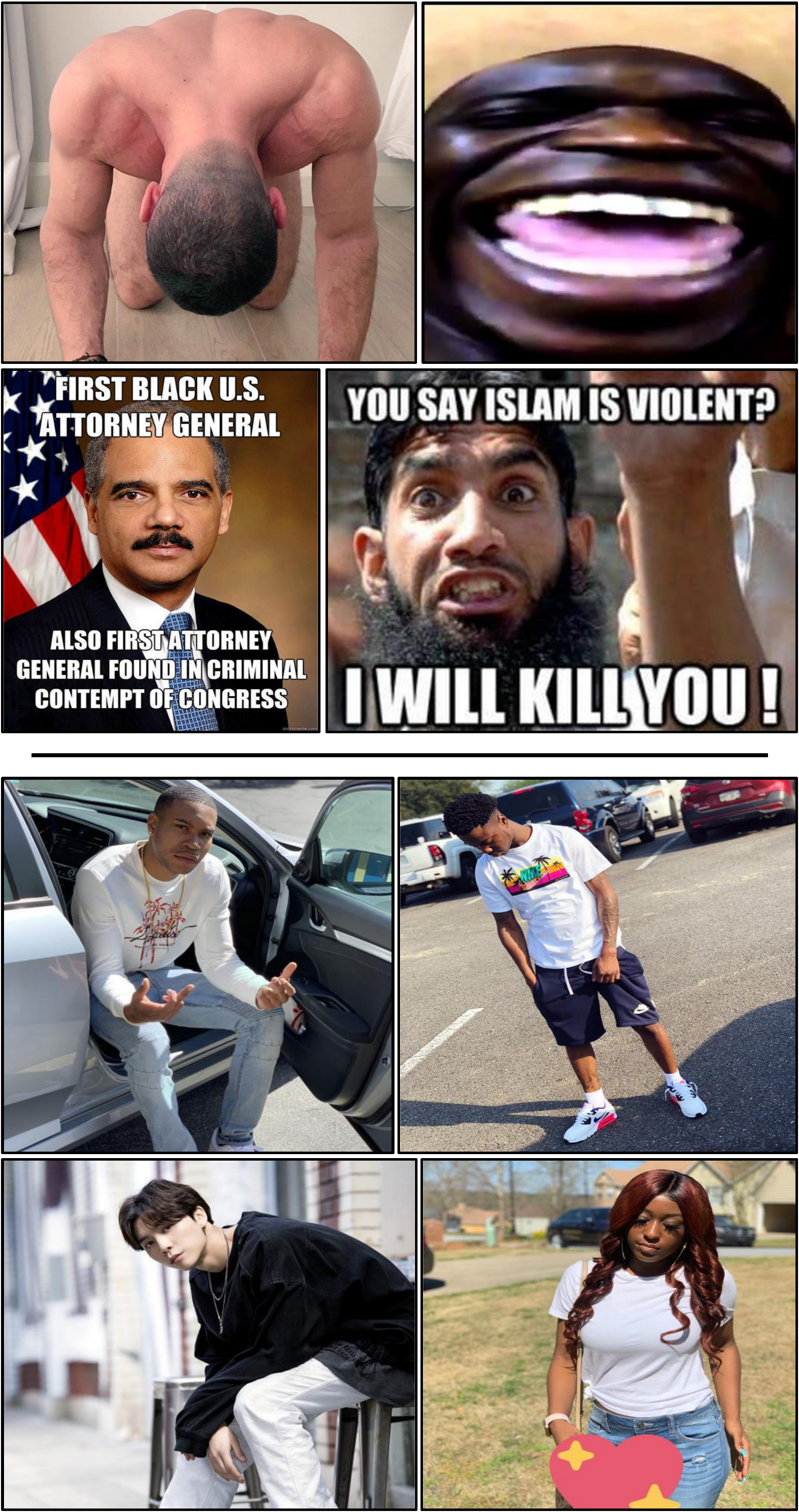}
   \caption{Top scored examples for hate (top) and for not hate (bottom) for the FCM model trained only with images.}
   \label{fig:FCM_I_results}
\end{figure}

% The FCM(I) model
The FCM trained only with images gets decent results, considering that in many publications the images might not give any useful information for the task. Fig.~\ref{fig:FCM_I_results} shows some representative examples of the top hate and not hate scored images of this model. Many hate tweets are accompanied by demeaning nudity images, being sexist or homophobic. Other racist tweets are accompanied by images caricaturing black people. Finally, MEMES are also typically used in hate speech publications. The top scored images for not hate are portraits of people belonging to minorities. This is due to the use of slur inside these communities without an offensive intention, such as the word \textit{nigga} inside the afro-american community or the word \textit{dyke} inside the lesbian community. These results show that images can be effectively used to discriminate between offensive and non-offensive uses of those words.

% Despite the image model learns useful image patterns, the multimodal models do not
Despite the model trained only with images proves that they are useful for hate speech detection, the proposed multimodal models are not able to improve the detection compared to the textual models. Besides the different architectures, we have tried different training strategies, such as initializing the CNN weights with a model already trained solely with MMHS150K images or using dropout to force the multimodal models to use the visual information. Eventually, though, these models end up using almost only the text input for the prediction and producing very similar results to those of the textual models. 
% Challenges of the task: Noisy annotations, small set of multimodal examples, complexity and diversity fo the relations. That's why the MM models models that work well in other tasks do nto succed in thsi application
The proposed multimodal models, such as TKM, have shown good performance in other tasks, such as VQA. Next, we analyze why they do not perform well in this task and with this data:

\begin{itemize}[noitemsep,leftmargin=*]
\item \textbf{Noisy data.} A major challenge of this task is the discrepancy between annotations due to subjective judgement. Although this affects also detection using only text, its repercussion is bigger in more complex tasks, such as detection using images or multimodal detection. 

\item \textbf{Complexity and diversity of multimodal relations.} Hate speech multimodal publications employ a lot of background knowledge which makes the relations between visual and textual elements they use very complex and diverse, and therefore difficult to learn by a neural network. 

\item \textbf{Small set of multimodal examples.} Fig.~\ref{fig:visual_hate_examples} shows some of the challenging multimodal hate examples that we aimed to detect. But although we have collected a big dataset of $150K$ tweets, the subset of multimodal hate there is still too small to learn the complex multimodal relations needed to identify multimodal hate.
\end{itemize}

\section{Conclusions}
In this work we have explored the task of hate speech detection on multimodal publications. We have created MMHS150K, to our knowledge the biggest available hate speech dataset, and the first one composed of multimodal data, namely tweets formed by image and text. 
We have trained different textual, visual and multimodal models with that data, and found out that, despite the fact that images are useful for hate speech detection, the multimodal models do not outperform the textual models. Finally, we have analyzed the challenges of the proposed task and dataset. Given that most of the content in Social Media nowadays is multimodal, we truly believe on the importance of pushing forward this research.
The code used in this work is available in \footnote{\url{https://github.com/gombru/multi-modal-hate-speech}}.

{\small
\bibliographystyle{ieee}
\bibliography{MyCollection}

\begin{thebibliography}{10}\itemsep=-1pt

\bibitem{Bazazian2017}
D.~Bazazian, R.~G{\'{o}}mez, A.~Nicolaou, L.~G{\'{o}}mez, D.~Karatzas, and
  A.~D. Bagdanov.
\newblock {FAST: Facilitated and Accurate Scene Text Proposals through FCN
  Guided Pruning}.
\newblock {\em Pattern Recognit. Lett.}, 2017.

\bibitem{Bazazian2016}
D.~Bazazian, R.~Gomez, A.~Nicolaou, L.~Gomez, D.~Karatzas, and A.~D. Bagdanov.
\newblock {Improving Text Proposals for Scene Images with Fully Convolutional
  Networks}.
\newblock {\em DLPR ICPR Work.}, 2017.

\bibitem{Cho}
K.~Cho, B.~{Van Merri{\"{e}}nboer}, C.~Gulcehre, D.~Bahdanau, F.~Bougares,
  H.~Schwenk, and Y.~Bengio.
\newblock {Learning Phrase Representations using RNN Encoder-Decoder for
  Statistical Machine Translation}.
\newblock {\em EMNLP}, 2014.

\bibitem{Davidson2017}
T.~Davidson, D.~Warmsley, M.~Macy, and I.~Weber.
\newblock {Automated Hate Speech Detection and the Problem of Offensive
  Language}.
\newblock {\em ICWSM}, 2017.

\bibitem{ElSherief2018}
M.~ElSherief, S.~Nilizadeh, D.~Nguyen, G.~Vigna, and E.~Belding.
\newblock {Peer to Peer Hate: Hate Speech Instigators and Their Targets}.
\newblock {\em ICWSM}, 2018.

\bibitem{Fukui}
A.~Fukui, D.~H. Park, D.~Yang, A.~Rohrbach, T.~Darrell, and M.~Rohrbach.
\newblock {Multimodal Compact Bilinear Pooling for Visual Question Answering
  and Visual Grounding}.
\newblock {\em EMNLP}, jun 2016.

\bibitem{Gao2018}
P.~Gao, P.~Lu, H.~Li, S.~Li, Y.~Li, S.~Hoi, and X.~Wang.
\newblock {Question-Guided Hybrid Convolution for Visual Question Answering}.
\newblock {\em ECCV}, 2018.

\bibitem{Gomez2018}
R.~Gomez, L.~Gomez, J.~Gibert, and D.~Karatzas.
\newblock {Learning to Learn from Web Data through Deep Semantic Embeddings}.
\newblock {\em ECCV 2018, MULA Work.}, 2018.

\bibitem{DianeLarlus2017}
A.~Gordo and D.~Larlus.
\newblock {Beyond Instance-Level Image Retrieval: Leveraging Captions to Learn
  a Global Visual Representation for Semantic Retrieval}.
\newblock {\em CVPR}, 2017.

\bibitem{Herz2012}
M.~Herz and P.~Molnar.
\newblock {\em {The content and context of hate speech: Rethinking regulation
  and responses}}.
\newblock Cambridge University Press, Cambridge, 2012.

\bibitem{Hosseinmardi}
H.~Hosseinmardi, S.~A. Mattson, R.~I. Rafiq, R.~Han, Q.~Lv, and S.~Mishra.
\newblock {Analyzing labeled cyberbullying incidents on the instagram social
  network}.
\newblock {\em Lect. Notes Comput. Sci.}, 9471:49--66, 2015.

\bibitem{Deng}
D.~Jia, D.~Wei, S.~R, L.~Li-Jia, L.~Kai, and F.-F. Li.
\newblock {ImageNet: A large-scale hierarchical image database}.
\newblock In {\em CVPR}, 2009.

\bibitem{Jiang2018}
W.~Jiang, L.~Ma, Y.-G. Jiang, W.~Liu, and T.~Zhang.
\newblock {Recurrent Fusion Network for Image Captioning}.
\newblock {\em ECCV}, 2018.

\bibitem{Karpathy}
A.~Karpathy and L.~Fei-Fei.
\newblock {Deep Visual-Semantic Alignments for Generating Image Descriptions}.
\newblock {\em IEEE Trans. Pattern Anal. Mach. Intell.}, 39(4):664--676, 2017.

\bibitem{Malmasi2017}
S.~Malmasi and M.~Zampieri.
\newblock {Detecting Hate Speech in Social Media}.
\newblock {\em Proc. Int. Conf. Recent Adv. Nat. Lang. Process.}, 2017.

\bibitem{Margffoy-Tuay2018}
E.~Margffoy-Tuay, J.~C. P{\'{e}}rez, E.~Botero, and P.~Arbel{\'{a}}ez.
\newblock {Dynamic Multimodal Instance Segmentation guided by natural language
  queries}.
\newblock {\em ECCV}, jul 2018.

\bibitem{Mikolov2013}
T.~Mikolov, G.~Corrado, K.~Chen, and J.~Dean.
\newblock {Efficient Estimation of Word Representations in Vector Space}.
\newblock {\em ICLR}, 2013.

\bibitem{Park}
J.~H. Park and P.~Fung.
\newblock {One-step and Two-step Classification for Abusive Language Detection
  on Twitter}.
\newblock {\em ALW1 1st Work. Abus. Lang. Online}, 2017.

\bibitem{Pennington}
J.~Pennington, R.~Socher, and C.~Manning.
\newblock {Glove: Global Vectors for Word Representation}.
\newblock {\em EMNLP}, 2014.

\bibitem{Saleem2017}
H.~M. Saleem, K.~P. Dillon, S.~Benesch, and D.~Ruths.
\newblock {A Web of Hate: Tackling Hateful Speech in Online Social Spaces}.
\newblock {\em First Work. Text Anal. Cybersecurity Online Saf. Lr.}, 2017.

\bibitem{Schmidt}
A.~Schmidt and M.~Wiegand.
\newblock {A Survey on Hate Speech Detection using Natural Language
  Processing}.
\newblock {\em Proc. Fifth Int. Work. Nat. Lang. Process. Soc. Media}, 2017.

\bibitem{Szegedy2015}
C.~Szegedy, W.~Liu, Y.~Jia, P.~Sermanet, S.~Reed, D.~Anguelov, D.~Erhan,
  V.~Vanhoucke, and A.~Rabinovich.
\newblock {Going deeper with convolutions}.
\newblock {\em Proc. IEEE Comput. Soc. Conf. Comput. Vis. Pattern Recognit.},
  2015.

\bibitem{Szegedy}
C.~Szegedy, V.~Vanhoucke, S.~Ioffe, and J.~Shlens.
\newblock {Rethinking the Inception Architecture for Computer Vision}.
\newblock {\em CVPR}, 2016.

\bibitem{GoogleOCR}
J.~Walker, Y.~Fujii, and A.~C. Popat.
\newblock {A Web-Based OCR Service for Documents}.
\newblock {\em IAPR International Workshop on Document Analysis Systems}, 2016.

\bibitem{Waseem2016}
Z.~Waseem.
\newblock {Are You a Racist or Am I Seeing Things? Annotator Influence on Hate
  Speech Detection on Twitter}.
\newblock {\em Proc. First Work. NLP Comput. Soc. Sci.}, 2016.

\bibitem{WaseemFix}
Z.~Waseem and D.~Hovy.
\newblock {Hateful Symbols or Hateful People? Predictive Features for Hate
  Speech Detection on Twitter}.
\newblock {\em Proceedings of the NAACL Student Research Workshop}, 2016.

\bibitem{Zhang2018}
Z.~Zhang, D.~Robinson, and J.~Tepper.
\newblock {Detecting Hate Speech on Twitter Using a Convolution-GRU Based Deep
  Neural Network}.
\newblock {\em Lect. Notes Comput. Sci.}, 2018.

\bibitem{Zhong}
H.~Zhong, H.~Li, A.~Squicciarini, S.~Rajtmajer, C.~Griffin, D.~Miller, and
  C.~Caragea.
\newblock {Content-driven detection of cyberbullying on the instagram social
  network}.
\newblock {\em JCAI Int. Jt. Conf. Artif. Intell.}, 2016.

\bibitem{Zhou2015}
B.~Zhou, Y.~Tian, S.~Sukhbaatar, A.~Szlam, and R.~Fergus.
\newblock {Simple Baseline for Visual Question Answering}.
\newblock {\em arXiv}, 2015.

\end{thebibliography}
}

\end{document}